\documentclass[10pt]{fairmeta}
\usepackage{fontspec}
\usepackage{graphicx}
\usepackage{caption}
\usepackage[numbers,sort&compress]{natbib}
\usepackage{amsmath,amssymb,booktabs,tabularx,array}
\usepackage{sectsty}
\usepackage[hidelinks]{hyperref}
\newfontfamily\calmregular{Outfit-Regular.ttf}
\newcommand{\calmfont}[1]{{\calmregular #1}}
\allsectionsfont{\calmfont}
\newcolumntype{L}{>{\raggedright\arraybackslash}X}
\title{\calmfont{HoloBoxing: A Simple Recipe for Humanoid Boxing with 7~Minutes of Single-GPU Training}}
\author[*]{\calmfont{Xihan Ma}}
\author[*]{\calmfont{Bo Zhang}}
\author[]{\calmfont{Boxiong Li}}
\author[]{\calmfont{Yuhan Bao}}
\author[]{\calmfont{Ziyue Liu}}
\author[\dagger]{\calmfont{Yucheng Wang}}
\contribution[*]{Equal contribution.}
\correspondence{Yucheng Wang}
\code{\url{https://github.com/HorizonRobotics/HoloMotion/tree/master/holoboxing}}
\date{23 October 2026}

\abstract{HoloBoxing presents a simple recipe for learning humanoid boxing by building on a pretrained whole-body controller. The recipe separates reusable motor skills from task decisions: a compact planner requests future body-point targets, while the frozen controller handles joint-level execution. A learned decoder maps 32-dimensional actions to structured target windows, and measured-state imitation initializes the planner before reinforcement learning. Task learning then refines interaction behavior while reusing the decoder and controller, avoiding the need to relearn whole-body control. This combination keeps task optimization focused on a compact upper policy and supplies an initial behavior before reward-driven exploration. Starting from that available controller, the recipe accounts for about seven minutes of additional training, including reused representation fitting and a single-GPU task run. This excludes controller pretraining, data preparation, evaluation, and search. Simulated boxing experiments and component controls evaluate the resulting behavior and the roles of imitation initialization and the target representation.}

\begin{document}
\maketitle
% BEGIN sections/frontpage_figures.tex
\noindent
\begin{minipage}{\linewidth}
\centering
\includegraphics[width=\linewidth]{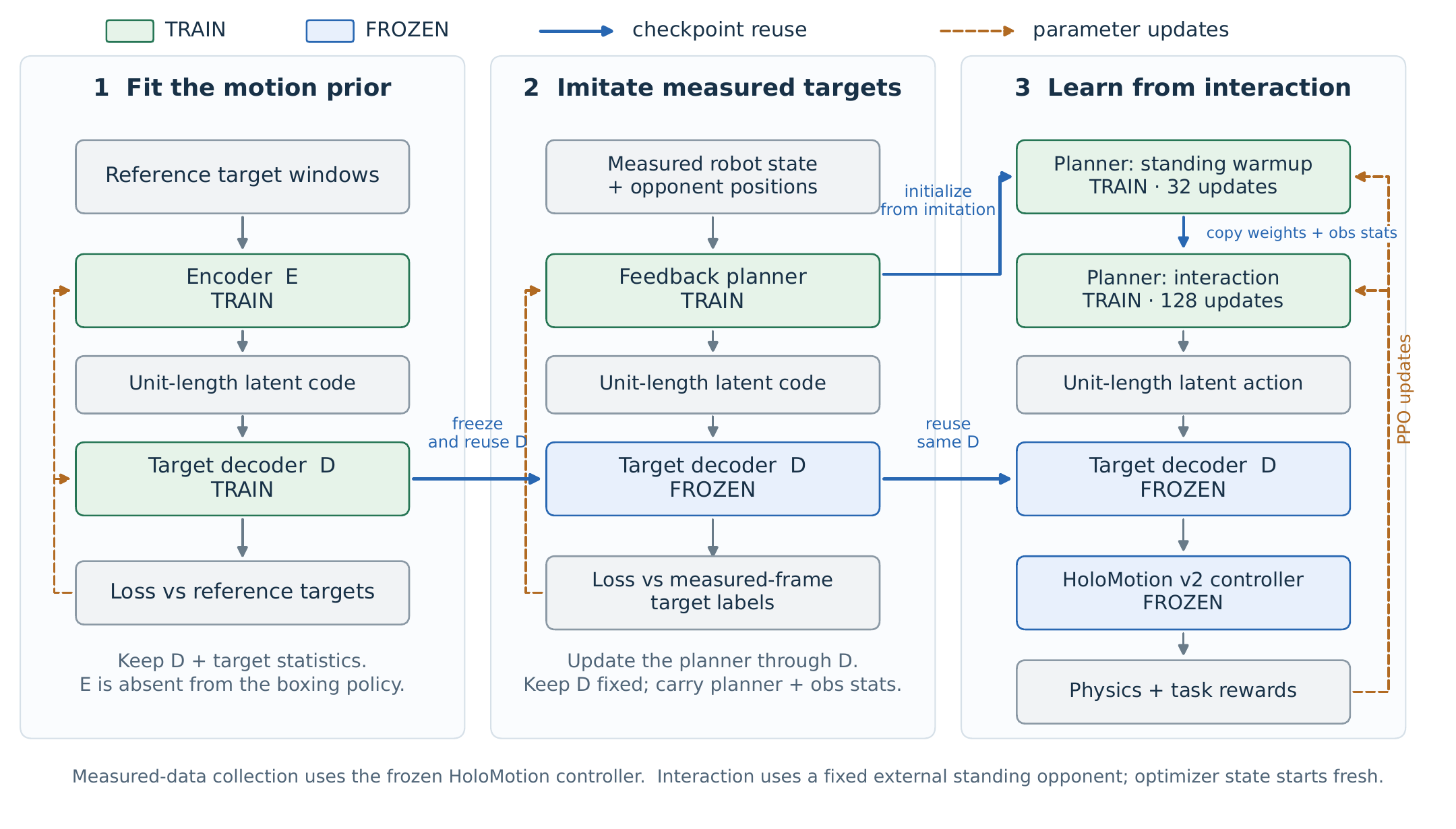}
\captionsetup{font=small,skip=3pt}
\captionof{figure}{What trains, what stays fixed, and what passes to the next stage. Blue arrows copy learned weights and statistics; gray arrows carry data or executed targets. Frozen decoder weights still allow gradients to reach the planner during imitation. Task reinforcement learning updates the planner from rollouts, without differentiating through the whole-body controller or physics.}
\label{fig:method}
\end{minipage}
% END sections/frontpage_figures.tex
\clearpage
% BEGIN sections/introduction.tex
\section{Introduction}
\label{sec:introduction}

A motion controller can reproduce a punch without deciding when or where to use it.
Humanoid boxing adds an opponent, changing contact, and the need to remain upright after an exchange.
Learning all of these behaviors from joint commands couples motor execution with task decisions.
HoloBoxing instead asks how far a simple, short training recipe can go when whole-body motor capability is already available.

Reusable whole-body controllers make task adaptation above existing motor capability a practical starting point.
The recipe separates what must be learned for boxing from what can be reused.
A pretrained HoloMotion v2 controller executes movement requests and remains frozen.
A learned target decoder represents short windows of future body-point poses with a 32-dimensional action.
A compact feedback planner selects these actions from measured robot state and opponent positions.
Task learning therefore adapts movement requests while retaining the underlying execution capability.

We initialize this planner through measured-state imitation before refining it with reinforcement learning.
The fixed controller first tracks reference motions to collect actual robot observations and future-target labels.
Supervised learning turns those pairs into an initial feedback policy.
Standing warmup and interaction against a fixed standing opponent then adjust behavior through rewards.
The decoder and lower controller are reused throughout, as illustrated in Figure~\ref{fig:method}.

These choices make the training problem smaller in three concrete ways.
Controller reuse removes joint-level motor learning from task adaptation.
The target representation reduces the planner's action dimension from 225 coordinates to 32 latent variables.
Imitation supplies an initial behavior before reward-driven exploration.
Together, they define the simple recipe whose training cost and resulting behavior we evaluate.

The primary recipe accounts for 7.0 minutes, combining a new single-RTX-4090 run with previously recorded target-representation fitting. A 21.4-minute long version provides a higher-budget comparison.
Simulation experiments assess whether this short adaptation produces interactive boxing, while matched-budget controls remove imitation initialization or the target decoder.
We also test observation choices and representation quality to explain the interfaces used by the recipe.
Detailed timings and competitive outcomes appear in the experiments.
% END sections/introduction.tex
% BEGIN sections/background.tex
\section{Background and motivation}
\label{sec:background}

Humanoid boxing combines motor execution with decisions about approaching, striking, and responding to an opponent.
Our focus is the additional learning needed when a general-purpose controller already supplies whole-body movement.
Three established ideas motivate a simple task-adaptation recipe.

\paragraph{Reuse motor capability.}
Motion imitation learns physically simulated movement from reference trajectories~\citep{peng2018deepmimic,luo2023phc}.
Recent humanoid systems extend this approach toward reusable whole-body control~\citep{fu2024humanplus,luo2025sonic,chen2026holomotion1}.
A task policy can request movement while a fixed controller converts those requests and robot feedback into joint commands.
HoloBoxing uses the released HoloMotion v2 controller in this role.
The boxing planner therefore learns task decisions above existing motor capability, rather than relearning joint-level execution.
This separation also defines the cost boundary: we measure new task training and exclude the inherited controller's pretraining.
The task-specific problem is consequently to choose useful movement requests in response to an opponent.
This differs from acquiring a broad motor repertoire and motivates evaluating adaptation time separately from controller development.

\paragraph{Represent movement requests compactly.}
Latent representations organize reusable movements and skills~\citep{peng2022ase,tessler2023calm,zhu2023ncp}, including spherical representations for state-conditioned motor control~\citep{tan2026slmp}.
HoloBoxing learns a decoder from a normalized 32-dimensional action to 225 future body-point target coordinates.
This gives the planner a compact action interface while retaining the same lower controller.
The representation supplies target structure, while the controller supplies motor feedback.
We evaluate both target error and tracking behavior.
The decoder and controller play different roles: one structures future requests, while the other responds to the robot's current physical state.
Keeping these roles separate lets the task policy reuse both interfaces as learning progresses.

\paragraph{Initialize before learning interaction.}
Imitation supplies an initial behavior, and reinforcement learning adjusts it through executed interaction.
Sports and boxing systems combine motion knowledge with task objectives in several forms~\citep{ren2025goalkeeper,chen2026badminton,zhang2026tennis,wang2026humanx,luo2024smplolympics,yin2026latentgames}.
Adapting frozen motor models also provides a reusable basis for new behaviors~\citep{vainshtein2026tasktokens,park2026maskadapt}.
HoloBoxing initializes its planner from measured robot observations collected during reference tracking, then refines it through standing warmup and interaction rewards.
The decoder and whole-body controller remain frozen throughout task learning.
Measured observations matter because the executing robot can deviate from the reference trajectory.
Training the planner on these states connects its initial behavior to the feedback available during interaction.
Reinforcement learning can then focus on adapting that behavior to task rewards.

Together, these choices confine task adaptation to a compact feedback planner with an initial behavior and a reusable execution interface.
The experiments measure the training cost and boxing behavior of this combined recipe.
% END sections/background.tex
% BEGIN sections/method.tex
\section{HoloBoxing}
\label{sec:method}

The recipe learns task decisions above a fixed motor controller.
It fits a target representation, initializes a feedback planner, and refines that planner through interaction while preserving whole-body execution.
For masked observation $M(o_t)$, the execution chain is
\begin{equation}
 z_t=\mathcal{N}(f_\theta(M(o_t))),\qquad
 g_t=D_\phi(z_t),\qquad a_t=\pi_{\mathrm{HM}}(c_t,g_t).
\label{eq:hierarchy}
\end{equation}
Here, $\mathcal{N}(u)=u/\|u\|$ uses a fixed fallback at zero.
The planner $f_\theta$ requests targets $g_t$ through decoder $D_\phi$; the frozen controller $\pi_{\mathrm{HM}}$ combines them with robot feedback $c_t$.

The planner has hidden widths 256 and 128 and predicts a 32-dimensional action.
Its 1095-dimensional normalized input retains 63 current-self features and 15 opponent-position features; the remaining entries are masked to zero.
Masking selects information without reducing the input-layer width.
The decoder expands the action to 225 position and orientation coordinates for five body points at five future times.
Thus, task learning operates through a small upper network and action interface while reusing motor execution.

\subsection{Compress movement requests before task learning}
The first stage fits an encoder $E_\psi$ and decoder $D_\phi$ to reference target windows $x$.
Using training statistics $(\mu_x,\sigma_x)$ and $\tilde x=(x-\mu_x)\oslash\sigma_x$, it minimizes
\begin{align}
 \hat x&=D_\phi(\mathcal{N}(E_\psi(\tilde x))),\\
 \mathcal{L}_{\mathrm{prior}}&=\mathbb{E}_{x}\left[\frac{1}{225}\|(\hat x-x)\oslash\sigma_x\|_2^2\right].
\label{eq:priorloss}
\end{align}
The symbol $\oslash$ denotes componentwise division.
The decoder restores physical target statistics and projects orientation features into valid rotations.
After validation selects a checkpoint, its weights and statistics remain fixed.
Only the decoder enters the executed boxing policy; the encoder remains an offline fitting and diagnostic component.
This replaces direct target prediction with a learned 32-dimensional action space.

\subsection{Initialize feedback before reward-driven exploration}
The frozen controller tracks reference motions to collect measured robot observations and future-target labels.
The planner learns from these pairs through the frozen decoder:
\begin{equation}
 \mathcal{L}_{\mathrm{IL}}(\theta)
 =\mathbb{E}_{(o,x)}[\ell_{\mathrm{target}}(D_\phi(\mathcal{N}(f_\theta(M(o)))),x)].
\label{eq:imitationloss}
\end{equation}
The target loss equally weights dimension-normalized position and orientation errors.
Gradients pass through the decoder to update the planner, while decoder weights remain fixed.
Reference windows use their source pelvis frame; collected labels use the measured robot's current pelvis frame with a shared horizontal alignment.
This expresses supervision in the feedback frame from which the controller must execute it.
Future reference states supply labels rather than actor inputs.
The selected planner weights and observation statistics initialize task learning, providing behavior before reward-driven exploration.
Appendix~\ref{app:method} specifies coordinate alignment and the complete target loss.

\subsection{Refine task decisions while retaining motor execution}
Proximal policy optimization (PPO)~\citep{schulman2017ppo} first runs 32 standing-warmup updates against a passive opponent.
The resulting weights and observation statistics initialize 128 interaction updates with fresh optimization state.
During interaction training, each learner uses a frozen copy of its own standing-warmup checkpoint as the opponent, without opponent refresh.
Its reward combines task shaping and adversarial motion style~\citep{peng2021amp} with weights 0.8 and 0.2.
Task terms guide approach, facing, contact, and balance; the match referee separately determines competitive outcomes.

The planner, value estimator, and style discriminator train during task learning.
The target decoder and HoloMotion controller stay frozen.
The value estimator predicts future reward, and the discriminator compares simulated motion with reference examples.
Reinforcement learning therefore refines task decisions while retaining the target representation and motor capability.
% END sections/method.tex
% BEGIN sections/evaluation_system.tex
\section{Evaluating short task adaptation}
\label{sec:infrastructure}

The evaluation asks whether the reused hierarchy produces interactive boxing after the recorded short training procedure.
It measures training time, competitive behavior, and the effects of removing task-policy components.
These measurements address different parts of the recipe: time quantifies adaptation cost, while behavior tests what that budget achieves.

All evaluated policies share the packaged robot, actuator settings, arena, simulator, and referee.
Each policy retains its original observation interface.
A valid hand contact scores a point, while a sustained fall can end the bout with a knockout.
We retain contacts and falls alongside match outcomes because an opponent can lose without receiving a scored punch.
Every bout lasts at most sixty seconds.

A fixed ten-round league gives each policy the same opponent coverage, with randomized sides and recorded initial-state seeds.
Elo summarizes outcomes within this frozen participant pool; component controls provide comparisons within the HoloBoxing recipe.
Recorded states, actions, and referee events allow replay without rerunning simulation.
Appendix~\ref{app:evaluation} specifies scoring thresholds, the rating equation, and replay conditions.
% END sections/evaluation_system.tex
% BEGIN sections/experiments.tex
\section{Experiments: training cost and resulting behavior}
\subsection{A small task learner above a fixed controller}
\label{sec:implementation}
The selected HoloBoxing recipe uses the Unitree G1 and the frozen HoloMotion v2 controller~\citep{holomotioncode}.
The motion corpus combines paired InterHuman data~\citep{liang2024intergen} and released RoboStriker trajectories~\citep{robostrikerpreview}.
The 14 original training motions total 212.88 seconds after retargeting; both participants and mirrored versions remain grouped within each split.
Prior fitting uses 21148 training windows, 11074 validation windows, and a separate test population of 3246 windows.
Measured-state imitation uses 2568 training examples from six InterHuman motions and 6448 validation examples from two different motions.
Appendix~\ref{app:implementation} provides source counts and training settings.

The decoder predicts five future poses for the pelvis, both wrists, and both ankles, spanning 20--100 ms.
The planner retains current robot state and opponent positions; time and score are unused.
PPO uses 128 parallel environments and 24-step rollouts.
Validation selects the reused prior. The primary recipe uses the last scheduled imitation and task-training checkpoints.
Appendices~\ref{app:implementation} and~\ref{app:method} retain dimensions, network widths, and optimization settings.

\subsection{Where the seven minutes are spent}
\begin{table}[ht]
\centering
\caption{Training-time accounting for HoloBoxing (7-min version). The new process uses one RTX 4090; target-prior fitting is reused from an earlier run.}
\label{tab:cost}
\small
\setlength{\tabcolsep}{4pt}
\begin{tabularx}{\linewidth}{LLr}
\toprule
Stage & Training or accounting scope & Runtime (s) \\
\midrule
Reused target prior & Earlier fitting run; epoch 148 selected & 47.2 \\
Imitation initialization & 20 epochs; final epoch selected & 6.7 \\
Standing warmup & 32 updates & 84.9 \\
Interaction learning & 128 updates & 275.3 \\
Other new-process time & Setup, saving, and exit & 8.1 \\
Total & Prior fitting plus complete new process & 422.2 \\
\bottomrule
\end{tabularx}
\end{table}
Table~\ref{tab:cost} accounts for 422.2 seconds, or 7.0 minutes.
The new process takes 375.0 seconds; previously recorded prior fitting adds 47.2 seconds.
Representation fitting and imitation together account for 53.9 seconds; standing and interaction account for 360.2 seconds.
The task learner reuses the lower controller throughout.

The total excludes controller pretraining, downloading, retargeting, observation collection, evaluation, and hyperparameter search.
The total measures the primary learner and target-representation fitting above these available components.
The primary version uses 128 interaction updates; the 21.4-minute long version uses 256 and a different recorded training path.

\subsection{Additional training above reusable motor capability}
The relevant cost is the additional training needed once motor control is available.
General-purpose whole-body controllers provide a reusable starting point; their original pretraining is outside this task-adaptation measurement.
We apply the same boundary to related systems by separating task learning from the acquisition of their inherited controllers and priors.

\begin{table}[ht]
\centering\small
\caption{Reported additional-training context. Existing motor-controller and prior pretraining are excluded from task-stage times. HoloBoxing additionally includes fitting its target representation and imitation initialization. Unreported denotes no wall-clock duration in the checked sources.}
\label{tab:reportedcost}
\begin{tabularx}{\linewidth}{lLL}
\toprule
Method & Existing motor capability & Reported additional training \\
\midrule
HoloBoxing & General-purpose HoloMotion v2 controller & 7.0 minutes including representation fitting, initialization, and task learning \\
SLMP & Tracking controller and latent prior & Approximately 8 hours for two-agent combat \\
RoboStriker & Tracking controller and latent prior & Task-stage wall-clock time unreported \\
GGSD & Joint skill discovery and gameplay learning & No separately timed post-controller adaptation stage reported \\
\bottomrule
\end{tabularx}
\end{table}

SLMP reports approximately eight hours for combat-policy training, excluding its tracking and latent-prior stages~\citep{tan2026slmp}.
Its hardware description lists a server with two RTX 4090 GPUs without specifying allocation to each stage.
RoboStriker reports a single RTX 4090 but no wall-clock duration in the checked paper or preview README~\citep{yin2026latentgames,robostrikerpreview}.
GGSD reports 70000 joint skill-learning and self-play updates, without a wall-clock duration or a separate comparable adaptation stage~\citep{rho2026ggsd}.
The evidence supports minute-scale boxing adaptation from an available general-purpose controller.
The reported training costs span different task objectives, hardware settings, observation interfaces, and inherited motor capabilities.

\subsection{Boxing behavior after short adaptation}

\label{sec:actualleague}

% BEGIN POOL PROTOCOL
Six frozen policies share the packaged robot, physical parameters, arena, and referee.
The participants are HoloBoxing (7-min version), its long version, two component removals, RoboStriker, and GGSD run2~\citep{ggsdcode}.
External opponents retain released weights without destination fine-tuning.
Each of ten rounds covers all fifteen unordered pairs, with randomized sides and recorded initial-state seeds.
Every pair therefore plays ten bouts, and each participant plays fifty.
Elo starts at 1500 with $K=16$; the fixed evaluation budget is independent of intermediate ranks.
Appendix~\ref{app:evaluation} gives the scoring thresholds and rating equation.
% END POOL PROTOCOL

\begin{figure}[ht]
\centering
\includegraphics[width=\linewidth]{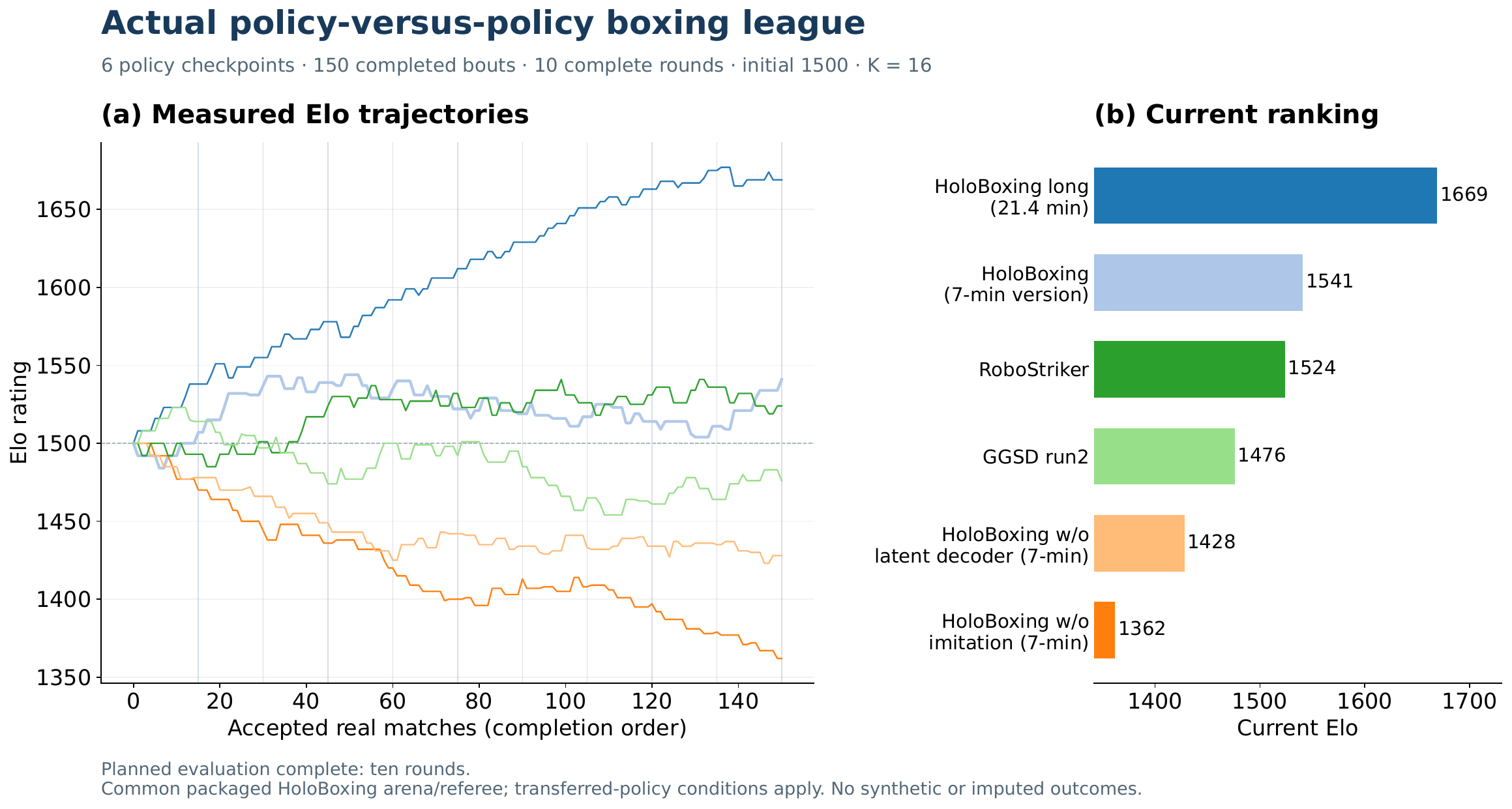}
\caption{Elo trajectories and final ranking over 150 simulated bouts. Each pair plays ten bouts under the same robot parameters, arena, and referee.}
\label{fig:actualelo}
\end{figure}

HoloBoxing (7-min version) ranks second in the fixed six-policy league; the long version ranks first.
Table~\ref{tab:leagueround} reports the complete outcomes, while Table~\ref{tab:leaguefamilies} separates opponent groups.
The primary version wins seven of ten bouts against GGSD and four of ten against RoboStriker.
Four of the primary version's wins contain no scored hand contact.

\begin{table}[ht]
\centering\small
\caption{Complete-round snapshot of the shared-arena league. Contact counts are candidate-relative totals. The 7-min version and both component removals share a 128-update final-stage budget. The long version uses 256 final updates. HoloBoxing variants are evaluated as distinct league participants.}
\label{tab:leagueround}
\begin{tabularx}{\linewidth}{Lrrr}
\toprule
Participant & Elo & W--L--D & Effective contacts \\
\midrule
HoloBoxing (long version, 21.4 min) & 1669 & 41--3--6 & 327:64 \\
HoloBoxing (7-min version) & 1541 & 26--17--7 & 76:196 \\
RoboStriker & 1524 & 28--22--0 & 337:29 \\
GGSD run2 & 1476 & 20--25--5 & 69:155 \\
w/o latent decoder (7-min version) & 1428 & 4--22--24 & 3:165 \\
w/o imitation (7-min version) & 1362 & 4--34--12 & 0:203 \\
\bottomrule
\end{tabularx}
\end{table}

% BEGIN LEAGUE FAMILY BREAKDOWN
Table~\ref{tab:leaguefamilies} separates external opponents from component controls.

\begin{table}[ht]
\centering\small
\caption{Primary 7-min-version results by opponent family in the same complete-round snapshot. Contacts and falls report the primary version first.}
\label{tab:leaguefamilies}
\begin{tabularx}{\linewidth}{Lrrr}
\toprule
Opponent family & W--L--D & Contacts & Falls \\
\midrule
GGSD & 7--3--0 & 21:16 & 0:5 \\
RoboStriker & 4--6--0 & 4:85 & 2:6 \\
Component removals & 13--0--7 & 30:0 & 0:0 \\
Long version & 2--8--0 & 21:95 & 0:0 \\
\bottomrule
\end{tabularx}
\end{table}
% END LEAGUE FAMILY BREAKDOWN

\subsection{Roles of initialization and the target representation}
The 7-min version and its component removals use one seed, 128 environments, and identical 24-step rollouts.
The primary HoloBoxing model provides the reference for the two component removals.
All complete 32 standing updates and 128 final updates, totaling 491520 learner transitions per group.
Matching applies to learner transitions; fitting costs differ when imitation is removed.
HoloBoxing (7-min version) retains all upper-policy components, with newly fitted imitation initialization and a reused frozen latent prior.
HoloBoxing w/o imitation initialization removes imitation parameter updates while preserving initial observation statistics and the same frozen prior.
HoloBoxing w/o latent decoder replaces the 32-dimensional latent action and decoder with a direct 225-dimensional target policy through the same frozen lower controller.
The 7-min version and direct-target variant each receive 20 imitation epochs from the identical measured training population.
For the primary version and each component removal, the final-stage opponent is a frozen copy of that run's own standing-warmup checkpoint, without opponent refresh.
The last scheduled checkpoint is selected before competitive evaluation.

\begin{table}[ht]
\centering\small
\caption{Single-seed component controls with 128 final updates. New-run minutes include startup, setup, training, saving, and exit. The shared prior is accounted for separately below.}
\label{tab:componentpilots}
\begin{tabularx}{\linewidth}{LLrr}
\toprule
Variant & Change from full recipe & New imitation epochs & New run (min) \\
\midrule
HoloBoxing (7-min version) & No component removed & 20 & 6.25 \\
w/o imitation (7-min version) & Remove imitation updates & 0 & 5.99 \\
w/o latent decoder (7-min version) & Predict 225 target values directly & 20 & 6.12 \\
\bottomrule
\end{tabularx}
\end{table}

The primary 7-min version combines a 6.25-minute new run with 0.79 minutes of previously recorded target-prior fitting, and trains against a frozen copy of its own standing-warmup checkpoint during final interaction learning. Each component removal likewise uses its own frozen warmup checkpoint as the final-stage opponent; the variant without imitation reuses the same target prior, whereas the direct-target variant has no target prior. The separately trained 21.4-minute long version uses a single RTX 4090 and 256 final interaction updates against its own frozen warmup opponent. The primary version and both component removals use 128 final updates.
Before training, the direct-target policy's exploration is calibrated against the latent policy using physical target perturbations, rather than equal numerical action noise.
The scale is selected from three prespecified candidates without match rewards.
Position and rotation perturbation magnitudes are comparable, while covariance and optimization geometry can still differ.
Appendix~\ref{app:additionalcontrols} reports the measured calibration values.
In the same league, the 7-min version ranks above both component removals and produces more scored contacts.
This is consistent with using imitation initialization and a structured target interface under the tested 128-update final-stage budget.
Removing the decoder also changes action dimension and exploration geometry, so its result concerns the complete action-interface choice.

\subsection{Choosing a compact but executable representation}
\paragraph{Select information before adding history.}
Table~\ref{tab:observations} compares eight observation choices with the same frozen decoder and three seeds per condition.
Current-state input with opponent information gives the lowest mean supervised validation loss, 16.9\% below ten-frame history with opponent input.
This supports the selected feedback information for imitation initialization.

\begin{table}[ht]
\centering
\caption{Planner observation study. Each condition uses three seeds and the same frozen decoder; values are selected validation losses.}
\label{tab:observations}
\small
\setlength{\tabcolsep}{4pt}
\begin{tabularx}{\linewidth}{LLLL}
\toprule
Self observation & Opponent input & Mean validation loss & Sample standard deviation \\
\midrule
Ten-frame history & Yes & 0.53513 & 0.00617 \\
Ten-frame history & No & 0.54799 & 0.01025 \\
Three-frame history & Yes & 0.46297 & 0.00333 \\
Three-frame history & No & 0.48215 & 0.00579 \\
Current state & Yes & 0.44447 & 0.00128 \\
Current state & No & 0.45782 & 0.00239 \\
Current pose & Yes & 0.47843 & 0.00285 \\
Current pose & No & 0.48925 & 0.00263 \\
\bottomrule
\end{tabularx}
\end{table}

\paragraph{Check execution as well as reconstruction.}
The selected reconstruction and uncompressed targets each complete 80 of 81 predefined validation windows, failing the same source window.
This supports retaining a compact representation without adding failures in that measured population.
A later candidate reduces reconstruction loss but introduces an additional tracking failure and loses required frame coverage in its separate fixed comparison.
It is therefore not promoted.
Appendix~\ref{app:tracking} specifies the populations, height gate, and coverage accounting.

Measured-frame labels are also harder to reconstruct than paired reference-frame targets with the same frozen representation.
This motivates checking the target interface under actual robot feedback, alongside reference reconstruction.
Appendix~\ref{app:design} gives the capacity, held-out, and paired-target diagnostics.
% END sections/experiments.tex
% BEGIN sections/discussion.tex
\section{Discussion}
\paragraph{Why the recipe is simple.}
HoloBoxing assigns motor execution, target structure, and task decisions to separate components.
The pretrained controller supplies feedback-based movement execution, while the decoder turns compact actions into requests that controller can follow.
Measured-state imitation connects robot observations to those requests before task rewards refine the planner.
The same interfaces persist across stages, so task adaptation reuses the decoder and controller instead of rebuilding their capabilities.

\paragraph{Why short adaptation is possible.}
The recipe confines task learning to the upper policy and its training objectives.
A 32-dimensional action replaces direct prediction of 225 future-target coordinates, and supervised examples provide the initial feedback behavior.
The recorded cost shows that representation fitting and initialization occupy a small part of the selected training path.
Most recorded training time is spent on standing and interaction, after reusable execution and target structure are already available.

\paragraph{What the experiments show.}
The seven-minute training path produces scored boxing interaction and ranks second in the evaluated pool. The 21.4-minute long version ranks first.
With 128 final-stage updates, the 7-min version yields stronger league outcomes than either component removal.
The representation studies also show why a low reconstruction loss alone is insufficient: the selected targets must remain physically executable.
Together, the experiments assess how the recipe's interfaces and initialization support the behavior obtained within its budget.

\paragraph{Limitations and future work.}
The study uses a small motion corpus and one task-training seed. The reported time measures additional learning above an available general-purpose controller, excluding its pretraining, preparation, and search. Independent seeds and matched timing studies are needed to isolate component effects on learning speed. Future work should broaden motion and reset coverage and test alternative lower controllers. The current experiments are simulation-only; real-robot boxing and transfer remain untested.

% END sections/discussion.tex
\section{Conclusion}
HoloBoxing combines a pretrained whole-body controller, a compact target representation, measured-state imitation, and task reinforcement learning.
Its simplicity comes from separating motor execution from task decisions and reusing the controller and decoder throughout task adaptation.
The latent target interface reduces the planner's action dimension, while imitation supplies an initial behavior before reward-driven interaction.
The primary recipe accounts for about seven minutes of additional training above an available controller. A 21.4-minute long version improves league outcomes at greater training cost.
In the evaluated six-policy league, the 7-min version ranks second and the 21.4-min version ranks first; at matched RL interaction budgets, the 7-min version also achieves higher Elo and more scored contacts than both component removals.

\section*{Acknowledgments}
We thank Lulu Cao and Yu Zheng for their time and effort in testing the HoloBoxing workflow end to end.
Their testing helped us check the consistency of the experimental pipeline across its stages.

\clearpage
\bibliographystyle{unsrtnat}
\bibliography{paper}
% BEGIN sections/appendix.tex
\clearpage
\appendix
\counterwithin{table}{section}
\counterwithin{figure}{section}
\section{Reproduction settings}
\label{app:implementation}
The selected recipe uses Unitree G1 and the frozen HoloMotion v2 controller~\citep{holomotioncode}.
Reference motions combine InterHuman~\citep{liang2024intergen} and released RoboStriker trajectories~\citep{robostrikerpreview}.
All frames, mirrored copies, participants, and overlapping windows from one original motion group stay in the same split.
The prior uses 21148 training, 11074 validation, and 3246 test windows.
Planner initialization uses 2568 measured training examples and 6448 validation examples from different motion groups.

\begin{table}[ht]
\centering\small
\caption{Motion coverage by source and split. Duration counts each original source motion once; mirrors and the two participants do not multiply it.}
\label{tab:datacoverage}
\begin{tabularx}{\linewidth}{LLrrrr}
\toprule
Source & Split & Groups & Files & Duration (s) & Prior targets \\
\midrule
RoboStriker & Train & 8 & 16 & 181.72 & 18092 \\
InterHuman & Train & 6 & 6 & 31.16 & 3056 \\
RoboStriker & Validation & 2 & 4 & 46.18 & 4598 \\
InterHuman & Validation & 2 & 2 & 64.96 & 6476 \\
RoboStriker & Test & 2 & 4 & 27.24 & 2704 \\
InterHuman & Test & 2 & 2 & 5.62 & 542 \\
\bottomrule
\end{tabularx}
\end{table}

\begin{table}[ht]
\centering\small
\caption{Implementation settings for the primary 7-min configuration.}
\label{tab:implementation}
\begin{tabularx}{\linewidth}{lLL}
\toprule
Symbol or component & Meaning & Value \\
\midrule
$d_o$ & Public observation dimension & 1097 \\
$K,H$ & Body points; future samples per point & 5; 5 \\
$d_p,d_r$ & Position and orientation feature dimensions & 3; 6 \\
$d_g,d_z,d_a$ & Target, latent, and motor-action dimensions & 225; 32; 29 \\
$\Delta t$ & Control interval & 20 ms; targets span 20--100 ms \\
Prior network & Encoder hidden widths & 512, 256; reversed in decoder \\
Task network & Planner hidden widths & 256, 128 \\
Style network & Discriminator hidden widths & 128, 64 \\
$N,T_r$ & Parallel environments; rollout steps & 128; 24 \\
$U_w,U_s$ & Standing and interaction updates & 32; 128 \\
Opponent & Own standing-warmup checkpoint (final stage) & Frozen; no refresh \\
$\sigma_0$ & Initial exploration scale & 0.1 standing; 0.2 self-play \\
$\rho$ & Exploration temporal correlation & 0.9 \\
$\alpha,\lambda_p$ & Task-reward and position-loss weights & 0.8; 0.5 \\
\bottomrule
\end{tabularx}
\end{table}

Prior fitting uses Adam, batch size 1024, and learning rate $10^{-3}$.
PPO uses five optimization epochs, four minibatches, adaptive learning-rate control initialized at $10^{-5}$, discount 0.99, advantage parameter 0.95, and clipping threshold 0.2.
Validation selects the reused prior; imitation uses 20 epochs with batch size 256 and learning rate $3\times10^{-4}$. The last scheduled imitation and interaction checkpoints are used.
The interaction stage collects 393216 learner transitions; standing adds 98304, totaling 491520.
The primary version and both component removals use seed 43. During final interaction, each run uses its own frozen standing-warmup checkpoint as the opponent, without refresh.
The 7-min total combines previously recorded prior fitting and the complete task-training process, each performed on a single RTX 4090. The 21.4-min long version was also trained on a single RTX 4090, with 256 final interaction updates compared with 128 in the 7-min version. Both totals exclude lower-controller pretraining, data preparation, evaluation, and hyperparameter search.

\section{Objectives and target coordinates}
\label{app:method}
Section~\ref{sec:method} defines the execution chain and the representation and imitation objectives.
This appendix specifies the coordinate alignment and detailed loss used with those interfaces.
For five points and five future samples, three position and six orientation features give $225=5\times5\times(3+6)$ target coordinates.

\paragraph{Measured-state imitation.}
The controller tracks source motions while collection records actual robot observations and future-target labels.
Future reference states provide labels, rather than actor inputs.
For measured pelvis pose $(c^m,R^m)$ and reference pose $(c^r,R^r)$, paired position targets satisfy
\begin{equation}
 p^m=(R^m)^{\mathsf T}(R^r p^r+c^r+\delta_t-c^m).
\end{equation}
The orientation encoding changes by $(R^m)^{\mathsf T}R^r$.
The shared alignment translation averages horizontal displacement errors across the $J$ participants:
\begin{equation}
 \delta_t=\operatorname{diag}(1,1,0)\frac{1}{J}\sum_{j=1}^{J}
 [(c^m_{t,j}-c^m_{t_0,j})-(c^r_{t,j}-c^r_{t_0,j})].
\end{equation}
Here, $t_0$ is the collection-window reset.
The same translation shifts all future points and participants, while orientations remain unchanged.
The target loss in Equation~\ref{eq:imitationloss} is
\begin{equation}
 \ell_{\mathrm{target}}(\hat x,x)=\frac{\lambda_p}{KHd_p}\|(\hat x^p-x^p)\oslash\sigma_p\|_2^2
 +\frac{1-\lambda_p}{KHd_r}\|(\hat x^r-x^r)\oslash\sigma_r\|_2^2.
\end{equation}
Training statistics use positive standard-deviation floors, and $\lambda_p=0.5$.
Gradients update the planner while preserving decoder weights.

\paragraph{Exploration and likelihood.}
PPO stores the raw Gaussian latent sample and its conditional log-probability before normalization.
Its likelihood ratio therefore concerns the sampled action rather than the normalized direction.
Standing warmup and task refinement use initial exploration scales 0.1 and 0.2, with temporal correlation 0.9.

\section{Representation checks and component controls}
\label{app:design}
\subsection{Observation and reconstruction studies}
Table~\ref{tab:observations} in the main text reports the observation study.
The remaining studies examine decoder capacity and target coverage.

The capacity study also uses three seeds per architecture.

\begin{table}[ht]
\centering\small
\caption{Prior-capacity study. Normalized error is dimensionless; its variation is the sample standard deviation across three seeds.}
\label{tab:priorcapacity}
\begin{tabularx}{\linewidth}{LLLL}
\toprule
Encoder hidden widths & Normalized error & Position error (cm) & Rotation error (degrees) \\
\midrule
256, 128 & $0.052182\pm0.001057$ & 2.653 & 5.249 \\
512, 256 & $0.046047\pm0.000709$ & 2.474 & 4.839 \\
512, 256, 128 & $0.055442\pm0.002174$ & 2.985 & 5.092 \\
\bottomrule
\end{tabularx}
\end{table}

\subsection{Physical tracking and held-out coverage}
\label{app:tracking}
The selected reconstruction and uncompressed targets each complete 80 of 81 validation windows, failing the same source window.
These windows cover six files from four original motion groups and 11014 fixed role-frames.
Each window lasts at most two seconds, following up to ten warmup frames.
Failure means pelvis or torso height below 0.30 m above the floor; source-reference height is retained for interpretation.
Windows from one motion are correlated observations.
A later candidate reduces validation loss from 0.045326 to 0.044545 but adds one tracking failure and misses 16 fixed comparison frames.
The accepted decoder therefore remains selected.

\label{app:diagnostics}
Held-out reconstruction is evaluated after selection, without reselecting the decoder.

\begin{table}[ht]
\centering\small
\caption{Offline reconstruction of the selected prior. These errors describe decoded reference targets before physical execution.}
\label{tab:priortest}
\setlength{\tabcolsep}{4pt}
\begin{tabularx}{\linewidth}{lLLLL}
\toprule
Split and source & Windows & Normalized error & Position (cm) & Rotation (degrees) \\
\midrule
Validation: RoboStriker & 4598 & 0.01985 & 1.595 & 3.441 \\
Validation: InterHuman & 6476 & 0.06342 & 2.902 & 5.890 \\
Test: RoboStriker & 2704 & 0.00658 & 0.985 & 2.269 \\
Test: InterHuman & 542 & 0.78406 & 6.752 & 19.055 \\
\bottomrule
\end{tabularx}
\end{table}

One seated-retreat InterHuman test clip contributes 97.7\% of that source's test reconstruction loss.
Both execution windows start below the upright-height threshold for the original and reconstructed targets, placing this seated motion outside the upright tracking evaluation.

\subsection{Reference-to-measured target transfer}
Paired source and measured labels use identical source frames and participants, with all representation weights and statistics fixed.
The comparison includes both the coordinate change and alignment translation defined above.

\begin{table}[ht]
\centering\small
\caption{Target-interface diagnostic on the measured validation population, with all representation weights and statistics frozen.}
\label{tab:measuredcompression}
\begin{tabularx}{\linewidth}{Lrrr}
\toprule
Target population & Samples & Position error (cm) & Rotation error (degrees) \\
\midrule
Paired source-frame references & 6448 & 2.90 & 5.88 \\
Measured-frame labels & 6448 & 6.30 & 10.36 \\
Measured subset, encoder initialization & 256 & 6.37 & 12.27 \\
Same subset, latent-only fitting & 256 & 6.03 & 8.61 \\
\bottomrule
\end{tabularx}
\end{table}

Latent-only fitting uses 128 evenly spaced examples per measured validation group, totaling 256 examples.
Only latent codes are optimized, with 100 Adam steps at learning rate 0.03.
The residual is an attained local-fit error, rather than a decoder expressiveness bound or initialized-planner result.

\subsection{Additional controls}
\label{app:additionalcontrols}
\label{app:lowerpilot}
Direct-target exploration is calibrated using three prespecified physical-noise candidates without match rewards.
At the final-stage setting, latent and direct-target policies induce position perturbations of 18.26 and 18.34 cm root-mean-square.
Their rotation perturbations are 0.713 and 0.701 radians.
The calibration yields comparable physical perturbation magnitudes while retaining the distinct latent and direct-target action representations.

\section{League rules and opponent interfaces}
\label{app:evaluation}
\label{app:leaguecontracts}
\label{app:audits}
All six current participants share the packaged robot, actuator parameters, arena, simulator, and contact referee.
Only the current ten-round, 150-bout league contributes to the reported ranking.

\paragraph{Scoring and ratings.}
Legal hand contacts against the opponent's head or torso, including pelvis targets, accumulate normal impulse at each 5 ms physics step.
A 5 N$\cdot$s threshold within 20 ms scores one point, consumes that impulse, and starts a 200 ms cooldown for that hand.
Ground contacts and foot strikes do not score.
Pelvis or torso height below 0.30 m above the ring starts a fall timer.
A continuous three-second fall confirms a knockout, awards the opponent 1000 points, and ends the bout.
Earlier observed falls determine knockout precedence; simultaneous confirmed falls produce a draw even when hit scores differ.
Otherwise, the 60-second timeout uses score, with equal scores producing a draw.

Each round covers all fifteen unordered pairs once, with randomized sides and recorded initial-state perturbations and sampling seeds.
Every participant starts at 1500 Elo, with pre-match ratings used for both updates:
\begin{equation}
 E_A=(1+10^{(R_B-R_A)/400})^{-1},\qquad
 \Delta_A=\operatorname{round}(16(S_A-E_A)).
\end{equation}
Here, $S_A\in\{0,\tfrac12,1\}$; A gains $\Delta_A$ and B loses the same amount.
Score margins and knockout bonuses do not multiply Elo updates.
Execution faults stop dispatch without replacement seeds, forfeits, or imputed outcomes.
The fixed ten-round budget is independent of intermediate ranks.

The two external opponents use the released RoboStriker G1 Stage-3 and GGSD run2 checkpoints.
Both retain source-trained weights and receive no destination fine-tuning.

\paragraph{Replay and measurement limits.}
Recorded robot states, actions, contacts, and referee events support replay without rerunning policies or physics.
A current dispatch check matches all 3000 non-timing frames across two RTX 4090 GPUs for one unchanged long-version--RoboStriker case.

\subsection{Inference timing}
\label{app:inference}
A fixed-pose RTX 4090 microbenchmark of the long version includes observation construction, planner, decoder, frozen controller, recurrent-state updates, and target conversion.
It excludes physics and output sanitation, and synchronizes the GPU around every timed interval.
After ten warmup iterations, 100 measured iterations average 3.182 ms, with empirical 99th percentile 3.373 ms.
The environment uses Warp 1.11.0, ONNX Runtime 1.22.0, CUDA 12.8.90, and driver 580.95.05.

\clearpage
\section{Recorded boxing examples}
\label{app:qualitative}
\begin{figure}[ht]
\centering
\includegraphics[width=\linewidth]{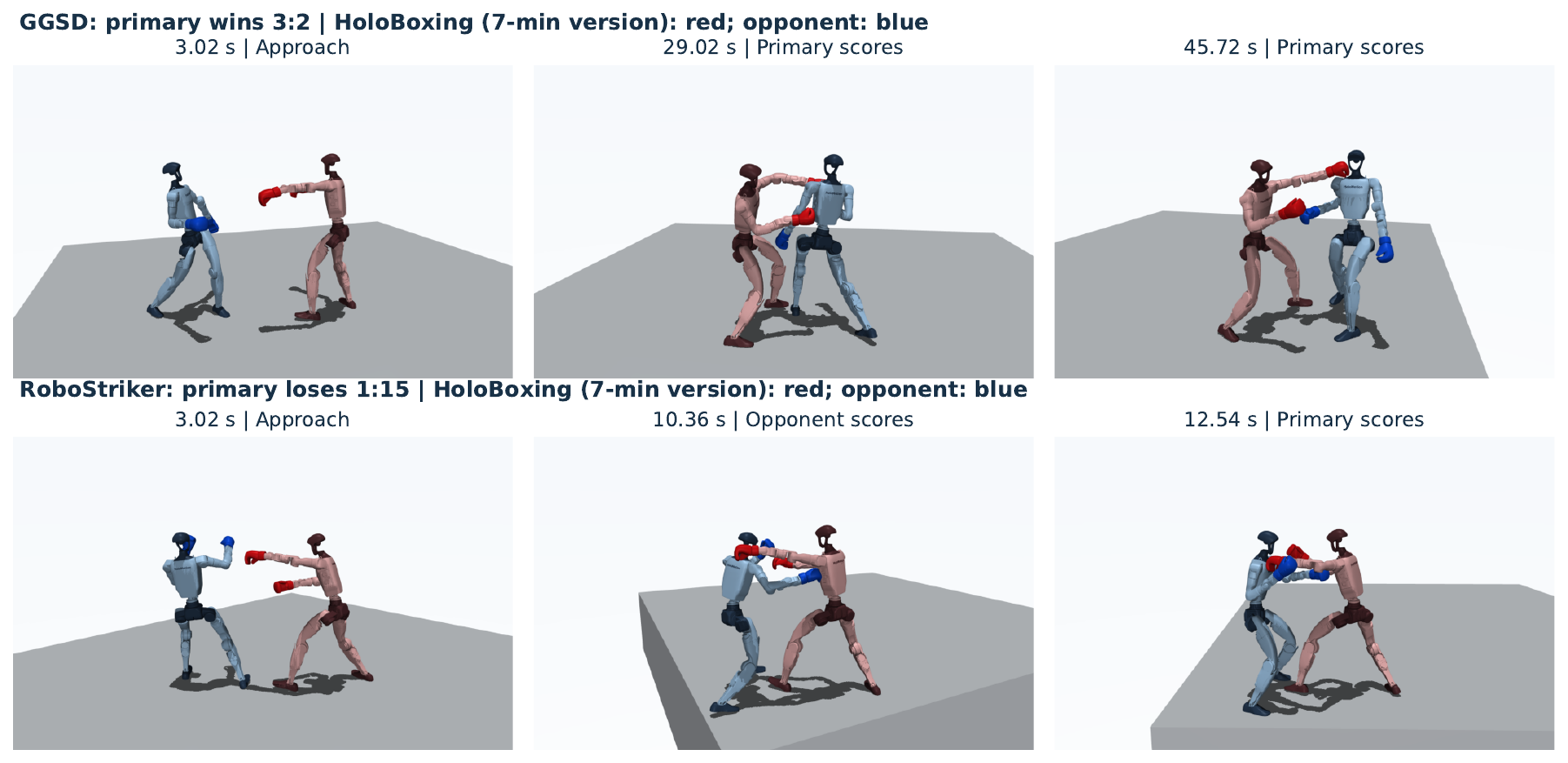}
\caption{Recorded boxing by HoloBoxing (7-min version, red) against GGSD (top, match r02-m005) and RoboStriker (bottom, match r05-m015). The opponent is blue. Top frames occur at 3.02, 29.02, and 45.72 seconds; bottom frames at 3.02, 10.36, and 12.54 seconds. Hit labels follow recorded referee events; the visible poses are the corresponding control-frame states. Camera and lighting are adjusted; wall faces and corner posts are hidden for visibility.}
\label{fig:sevenexamples}
\end{figure}

Figure~\ref{fig:sevenexamples} shows recorded bouts from the primary 7-min version.
The examples include a win against GGSD and a loss against RoboStriker, illustrating approach and scored exchanges in both directions.
Frames are reconstructed from saved robot poses without policy inference or physics stepping.

% END sections/appendix.tex
\end{document}